\documentclass[letterpaper, 10 pt, conference]{ieeeconf}  

\IEEEoverridecommandlockouts                              

\usepackage{graphicx} 
\usepackage{amsmath} 
\usepackage{hyperref}
\usepackage{subfig}

\makeatletter
\let\NAT@parse\undefined
\makeatother
\usepackage[numbers]{natbib}
\usepackage[font=small,labelfont=bf,tableposition=top]{caption}
\usepackage{color}
\usepackage{epstopdf}
\DeclareMathOperator*{\argmin}{arg\,min}

\title{\LARGE \bf
Addressing Challenging Place Recognition Tasks using Generative Adversarial Networks
}
\author{Yasir Latif, Ravi Garg,  Michael Milford and Ian Reid
\thanks{YL, RG, and IR are with the Australian Center for Robotic Vision (ACRV) at the University of Adelaide 
{\tt\small firstname.lastname@adelaide.edu.au} MM is with the ACRV at Queensland University of Technology, Brisbane.
{\tt\small michael.milford@qut.edu.au}.
Further details can be found at 
{\url{https://ylatif.github.io/projects/PRwithGANS/}}
}
}

\begin{document}

\makeatletter
\let\@oldmaketitle\@maketitle
\renewcommand{\@maketitle}{\@oldmaketitle
\centering \includegraphics[width =\textwidth]{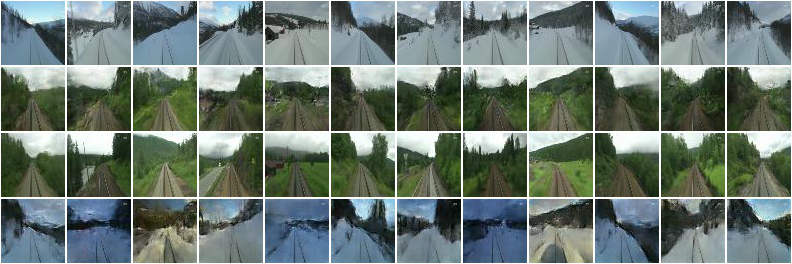}%
\captionof{figure}{Learned image translations: (\textbf{Rows 1,3}): Original images (\textbf{Rows 2,4}): Winter to summer
and summer to winter from the corresponding images in rows 1 and 3. Generative Adversarial Networks are able to map from
one domain to another with high visual fidelity. Discriminators in each domain learn features specific to the
domain that are useful for place recognition tasks. Instead of matching summer to winter (rows:1,3), which is a
difficult perception task, summer images can be reliably matched to generated-summer (rows:2,3).}%
\label{fig:main}%
\bigskip}
\makeatother

\maketitle
\thispagestyle{empty}
\pagestyle{empty}

\begin{abstract}
Place recognition is an essential component of Simultaneous Localization And Mapping (SLAM). 
Under severe appearance change, reliable place recognition is a difficult perception task 
since the same place is perceptually very different in the morning, at night, or over different seasons. 
This work addresses place recognition as a domain translation task. 
Using a pair of coupled Generative Adversarial Networks (GANs), 
we show that it is possible to generate the appearance of one domain (such as summer) from another (such as winter) 
\textit{without} requiring image-to-image correspondences across the domains.
Mapping between domains is learned from sets of images in each domain without knowing the instance-to-instance correspondence
by enforcing a cyclic consistency constraint.
In the process, meaningful feature spaces are learned for each domain, the distances in which can be used for the task of place recognition.
Experiments show that learned features correspond to visual similarity and can be effectively used for place recognition across seasons.
\end{abstract}

\section{Introduction}
The problem of SLAM has matured enough for it to move out of a laboratory setting and be applied to real world scenarios, 
where it is expected to operate reliably over the life span of the robot \cite{cadena2016past}. 
One of the essential components of any SLAM system is \textit{place recognition} or \textit{loop closure detection} 
which allows a mobile robot to reduce drift as well as uncertainty in its trajectory by detecting revisits.
While appearance change is small and gradual in the frame-to-frame timescale and poses no challenge for tracking and data-association,
over a long operational time, the appearance of a place can change significantly due to changes in illumination or weather conditions.
Place recognition in presence of severe appearance change is a difficult perception problem and is the subject of this work.

Changes in illumination can make images from the same place appear drastically different from each other.
If we can model the transformation from a source domain (such as summer) to a target domain (such a winter),
we can reliably recognize places originally observed in different domains by transforming them to a common domain.
%
This is so-called ``domain translation'' task \cite{ganin2016domain,tzeng2015simultaneous}, that is,
given an image in one domain (for example summer), generate the corresponding image in a target domain (for example winter),
such that the structure remains the same, but the transient conditions are varied. 

Recently, Generative Adversarial Networks (GANs) ~\cite{goodfellow2014} 
have been shown to generate realistic domain specific image.
We use a pair of coupled gains to generate realistic images in the target domain from images in the source domain and vice versa.
Domain translation normally requires images with known correspondences across the domains,
but labeling such correspondences is time consuming and might not always be possible.
In this work, the relationship between the two domains is learned from sets of images in each domain 
{\em without} requiring one-to-one image correspondences across the domains.
Images generated via domain translation can be used directly for place recognition using pixel-wise differences,
however, we show that the features spaces learned by the discriminator for each domain are more informative than pixel-wise differences, 
and therefore more useful for the place recognition task.
Some winter-summer translations using the proposed method are depicted in Fig. \ref{fig:main}.
The first and third rows contain real images, while the other two rows contain images generated from the corresponding
domain. If we have correctly modeled the domain transfer functions, images in row 1 and 2 should look the same as images in row 4 and 3 respectively.

The rest of the paper is organized as follows: in Section \ref{sec:relatedWork} we present an overview for different
approaches that have been devised for place recognition under extreme weather changes and present the relevant
literature for GANs as well. 
In Section \ref{sec:gansForPR}, we present the specific network used in this work and briefly justify why it is correct
choice for the task. Experiments are then presented that highlight the properties of the learned features and their
performance for the place recognition task. Finally, conclusions and future work is presented. 

\section{RELATED WORK}
\label{sec:relatedWork}
Visual place recognition is a well studied problem and a good overview can be found in \cite{PR_survey}.
Traditionally, it has been addressed using feature-based approaches, where
features descriptors extracted from an image serve are used as a representation for the place being observed \cite{cummins2008fab}\cite{galvez2012bags}.
Feature-based methods operate well in the short time frame where change in illumination and appearance is limited. 
They would, however, fail in situations where the change in the environment is greater than the invariances 
(such a rotation, scale,
illumination etc.) offered by the underlying feature descriptors. 

\subsection{Visual Place Recognition}

Visual place recognition can be broadly divided into two dominant approaches; 
\textbf{a)} feature-based methods that represent an image as a set of features descriptors extracted from interesting locations in the image, and 
\textbf{b)} image-based methods that try to reason using all the image information, instead of extracting features \cite{PR_survey}. 
Each approach has it own limitations: feature-based methods may fail under illumination changes while image-based methods are sensitive to view-point changes.

Several approaches have been introduced to address the problem of visual place recognition under extreme appearance changes.
These approaches can be further divided into two broad categories: 
\textbf{a)} methods that mitigate the effect of visual change by learning/utilizing illumination invariant descriptors, and 
\textbf{b)} methods that learn how to transform images from one visual domain to another. 
Belonging to the first category,
SeqSLAM \citep{seqSLAM} represents images as mean and variance normalized patches before reasoning about their similarity.
They show that even though a single image might not contain enough information, 
sequences of images, can lead to successful place recognition under extreme weather changes.
Along similar lines, Naseer, et al. \cite{naseer2014robust} formulate the problem of matching sequences as a network flow problem,
using HOG features over a grid in an image as the illumination invariant representation of the image. 
For traditional feature descriptors,
\cite{valgren2010sift} showed that U-SURF is more robust to illumination changes than SIFT
and SURF  which, together with epipolar constraints, can be used to close loops across different illumination conditions.
The effect of illumination change over the span on a day is studied in various works \cite{ross2014method,
ross2013novel} and showed that U-SIFT \cite{lowe1999object} has the greatest illumination invariance among traditional
feature descriptors. 

Instead of whole images or features, several structure based methods have also been proposed that utilize edges in the
images as the illumination invariant description \cite{eade2006edge,nuske2009robust}. Techniques such as shadow removal
\cite{corke2013dealing} have been used to remove unwanted artifacts from images, leading to improvement in the place recognition task.
Additionally, features from Convolutional Neural Networks (CNNs) have also successfully used for the task of place recognition
due to their invariance to illumination and viewpoint changes \cite{sunderhauf2015performance}.

This work takes a different approach towards the problem of place recognition.
We aim to generate the appearance of a place given the current environment conditions (for example winter) from an image captured under different conditions (for example summer).
We only assume that we have seen sets of images captured under both conditions, instead of image-to-image correspondences, 
which can be used to learn the transformation from one domain to the other.
Our work is closest in spirit to \cite{neubert2013appearance}, where the transformation is learned through Bag-of-words descriptors, however, 
we do not utilize paired correspondence across seasons at the learning stage.

\subsection{Generative Adversarial Networks}

GANs \cite{goodfellow2014} have been successful applied to the task of domain specific image generation. 
In its simplest form, a GAN consists of two components: 
a generator $\mathcal{G}$, which randomly samples from a latent space  and aims to generate a realistic image that resembles
images from a domain being learned (such as faces, cars, rooms, etc),
and a discriminator $D$ whose task is to correctly discriminate between real and generated (fake) images. 
These two components compete to outperform each other: 
the generator aims to produce images in such a way that the discriminator would classify them as real, 
and the discriminator aims to correctly identify which image is real and which is generated.
In a single pass through such a network, a random sample from a known distribution is converted to an image
by the generator, which, along with a real image from the domain, is shown to the discriminator for classification.
The classification error between the real and generated image serves as an update for the discriminator as well as the
generator, one trying to minimize it and the other trying to maximize it.
As a result of this adversarial training, the generator learns a mapping from the latent space to the image domain
under consideration. The latent space serves as a representation for the images in the domain which is shown to be
smooth (each point in the space correspondence to an image) and supports vector operations \cite{radford2015unsupervised}.

The common analogy of how GANs work is to consider it as a contest between a forger who produces fake banknotes
and a bank which aims to tell real and fake notes apart. 
They are both locked in an adversarial zero sum game: where success for one means a loss for the other. 
The better the bank gets at detecting the fake banknotes, the better the forger has to get to defeat it.
Similarly, the more adapt the forger gets, the better the bank has to be in order to detect the fakes banknotes.
Over time, they reach an equilibrium, in which the forger is the best it can be at producing fake banknotes, 
and the bank becomes the best it can be at detecting the forgeries. 

There is significant literature regarding GANs and their various applications,
however, we are particularly interested in the types of networks that perform domain adaptation/transfer.
One such work is DiscoGAN~\cite{DiscoGAN}, which discovers mappings between two image domains using unpaired
examples from both domains. They show that a pair of coupled GANs can discover relationships between styles of handbags
and shoes and discover common attributes (such as orientation) between images of cars and human faces, etc. 
Along similar lines, CycleGAN~\cite{CycleGAN2017} enforces cyclic-consistency to learn one-to-one mapping between domains and shows
translation from satellite images to maps. If the two domain have paired information, pix2pix~\cite{pix2pix2016},
 can be used to learn the translation from one domain to another.  %
The problem of domain adaptation  has also been considered in \cite{WulfmeierIROS2017} to improve the performance of free-space estimation under varying illumination conditions.

\section{Place Recognition using GANs}
\label{sec:gansForPR}
\begin{figure}
\centering
\includegraphics[width=\columnwidth]{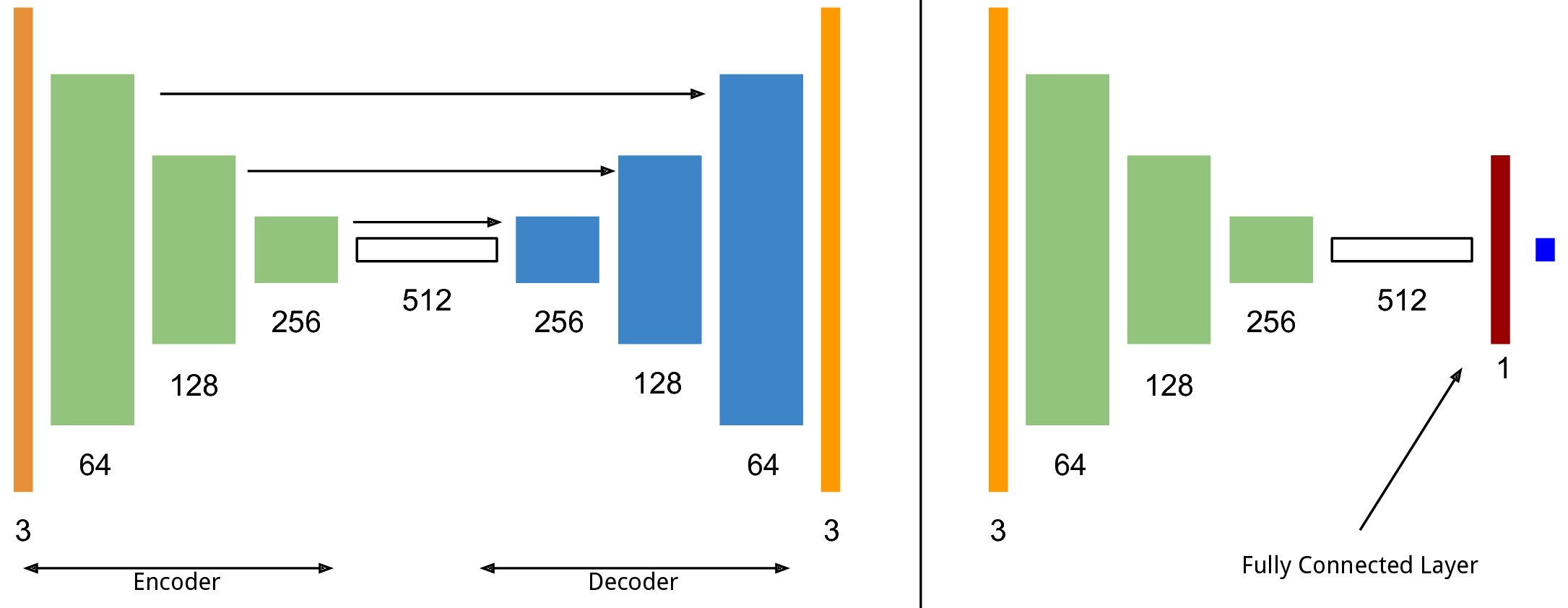}
\caption{Architecture for the Generator (left): 
Orange blocks represent input and output images of size 64$\times$64. 
Numbers below the block represent the number of channels output by the block. Each encoding block (green) consists of a
convolution layer (filter size 4, stride 2), followed by batch normalization and Leaky Rectified Linear Unit (slope 0.2), 
Each decoder block (blue) consists of deconvolution layer, batch normalization and Rectified Linear Units. 
Solid arrows indicate skip connection which are added to the output of the corresponding layers. 
The discriminator (right) has the same structure as the encoder and has a two fully connected layer at the end which is then mapped to a single output (dark blue) via sigmoid for classification. The first fully-connected layer is used as feature extractor for place recognition.
}
\label{fig:archi}
\vspace*{-5mm}
\end{figure}


\begin{figure*}
\centering
\includegraphics[width=0.6\textwidth]{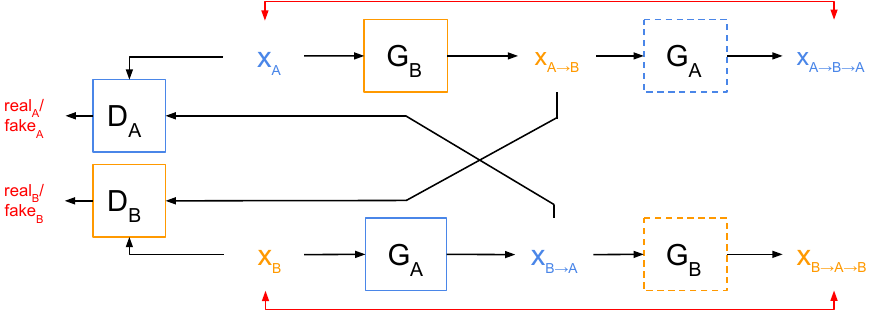}
\caption{
Coupled GANs for domain translation. Blue and orange represent different domains. 
$\mathbf{x}_A$ and $\mathbf{x}_B$ are images from domains A and B respectively. 
$\mathbf{G}_B$ takes an image from domain A and generates the corresponding image in domain B ($\mathbf{x}_{A \to B}$),
which is then compared to a real image from domain B by discriminator $\mathbf{D}_B$ to assess its quality.
Using the current state of the generator $\mathbf{G}_A$, it is also translated back to the original domain A and error is computed between the input image and the reconstruction.
Input image $\mathbf{x}_B$ follow a similar path.
Classification and reconstruction losses (in red) are used to train $\mathbf{G}_{A/B}$ and $\mathbf{D}_{A/B}$.
The fully connected layer in each discriminator is used as the encoding feature space for images in the domain. (See Fig. \ref{fig:archi})
}
\label{fig:system}
\end{figure*}

The holy grail for place recognition methods is to find a suitable representation space for images in which these properties
hold: 
\textbf{(a)} distance between images depends on the observed structure and not the appearance, 
\textbf{(b)} the representation is view-point invariant, or at least view-point aware, so that images that observe the
same structure from different viewpoints are placed close to each other, and
\textbf{(c)} the distances in this space are meaningful, that is, the larger the distance between two images in this space, the
less likely it is that they have common structure (in short, a vector space in which the triangular inequality holds
based on the structure being observed).
If such a representation is found, then place recognition simplifies to the nearest neighbour search because
all other distractors (such as illumination, view-point, etc) have been taken into consideration and normalized.
We show that by using GANs, something akin to such a representation can be learned.
We first highlight the useful properties of GANs which make them ideal for the task of place recognition
and then describe the architecture of the system used in this work. 

The task of a simple GAN is to capture the distribution of a given domain. 
This is achieved by training the generator ($\mathbf{G}$) to map random samples from a hidden space to images of the particular domain.
For the task at hand, we are interested in encoding an image into the latent space, 
and a normal GAN does not have the provision for it.
We need a generator that can take an input image, instead of a random sample, 
and generates an image at the output. 
Therefore, we use a simple encoder-decoder network in place of a generator, 
where the encoder part maps the image to an encoding space, and the decoder generate the output image.
Such a setup can then be used to map one domain to another by having different
inputs to the encoder-decoder (source domain) and discriminator (target domain). 

For the domain translation task, the source domain can be mapped to a target domain in infinitely many ways, 
since GANs maps one distribution to another.
To discover instance-level relationships, the problem can be constrained by using two coupled GANs,
that simultaneously translate from domain A to B and vice versa. 
This allows cyclic constraints \cite{CycleGAN2017, DiscoGAN}:
an image from domain A, after translation to domain B and retranslated back to domain A, 
should match the original image.
We provide a general overview of such an architecture below.
Assuming that there exists a generator function $\mathbf{G}_{B}$ that can map an image in domain A, $\mathbf{x}_A$ to an image in domain
B such that $\mathbf{x}_{A \to B} = G_B(\mathbf{x}_{A})$, where $\mathbf{x}_{A \to B}$ means an image from domain A translated to domain B.
Additionally, another generator  $\mathbf{G}_{A}$ maps in the reverse direction, that is, 
$\mathbf{x}_{B \to A} = \mathbf{G}_{A}(\mathbf{x}_{B})$. 
In order to translate between the two domains, we need to train these two generators. 
However, without any further restriction, there are infinitely many functions that can satisfy these
requirements. 
As proposed in \cite{DiscoGAN}, we use the cyclic consistency constraint to restrain the system to do one-to-one mapping between
the two domains by minimizing the cyclic reconstruction loss:

\begin{equation}
\argmin_{\mathbf{G}_A, \mathbf{G}_B} ||\mathbf{x}_A - \mathbf{x}_{A \to B \to A}||_2
\end{equation}
\noindent 
and
\begin{equation}
\argmin_{\mathbf{G}_A, \mathbf{G}_B} ||\mathbf{x}_B - \mathbf{x}_{B \to A \to B} ||_2
\end{equation}
\noindent
which encourages the two generators to be inverses of each other, 
that is $\mathbf{G}_{A}( \mathbf{G}_{B} ( \mathbf{x}_A )) \to \mathbf{x}_A $
and $\mathbf{G}_{B}( \mathbf{G}_{A} ( \mathbf{x}_B )) \to \mathbf{x}_B $.

Discriminators $\mathbf{D}_A$ and $\mathbf{D}_B$ work in each domain and try to discriminate between 
$\mathbf{x}_A$, $\mathbf{x}_{B \to A}$ and $\mathbf{x}_B$,$\mathbf{x}_{A \to B}$ respectively. 
In the process, the discriminators learn feature spaces specific to each domain. 

Fig. \ref{fig:system} provides an overview our set up. 
Each generator-discriminator pair works in a single domain:
translating from the corresponding domain and differentiating between real and fake images, respectively.
Each generated image is then fed into the other generator to get an image in the original domain, which forms the cyclic constraint.
The gray lines represent cyclic pixel-wise Mean Squared Error constraints. 

\subsection{Network architecture}
Fig. \ref{fig:archi} shows the architecture of the generator and discriminator networks.
The generator consists of a set of convolution and deconvolution blocks as shown in Fig. \ref{fig:archi}.
In order to preserve edge information, we use skip connections so that useful information about edges can
flow easily through the network. 
The bottleneck layer is the encoding space of the generator which is then used to
generate the image in the other domain.
The discriminator consists of the encoder part of the generator with an additional fully connected layer at the end which serves as the feature layer in the discriminator. 
For classification, this layer is mapped to a single output via a fully connected layer followed by a sigmoid activation function.
In the experimental section, we show that this layer learns informative features that can be used for place recognition.

%

\subsection{Place Recognition using GANs}

Consider the case where we have seen the same place in two different weather conditions so
that images from them can be used to train for the domain translation task.
Each domain (summer, winter) consists of a set of training images from which the relation between the domains can be learned.
We do not utilize image-to-image correspondence across domain as it is difficult to annotate images compared to collecting images of a particular domain.

At test time, we assume that the current conditions are known and we want to do close loops against images from the original domain.
We generate images in the other domain using the appropriate generator, 
which ideally would generate the exact image,
given that they are from the same viewpoint. 
Another point of comparison is the features learned by the discriminator.
We can go from $\mathbf{x}_A$ to $\mathbf{x}_{A \to B}$ and compare the features from $\mathbf{D}_B$ for the pair $\mathbf{x}_B$ and $\mathbf{x}_{A \to B}$, 
where the first image is from domain A, and second is an image from domain B translated to A. 

In the experiment section, we present results for place recognition in the feature space and explore its properties.
\begin{figure}
\centering
\includegraphics[width=.8\columnwidth]{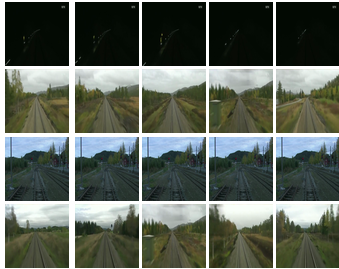}
\caption{Images close to each other in the feature space of summer. Reference Image (left) images sorted by distance
(low to high).}
\label{fig:similarImages}
\end{figure}
\begin{figure*}
\centering
\subfloat[Normalized Distances: (Left to right) distance in image space, zoomed-in to show details, distance in the feature space, zoomed-in to show details.
Each row and column is normalized by the value of the diagonal. Ideally all off-diagonal elements should have a large distance (more red is better).]{
\includegraphics[height=4cm]{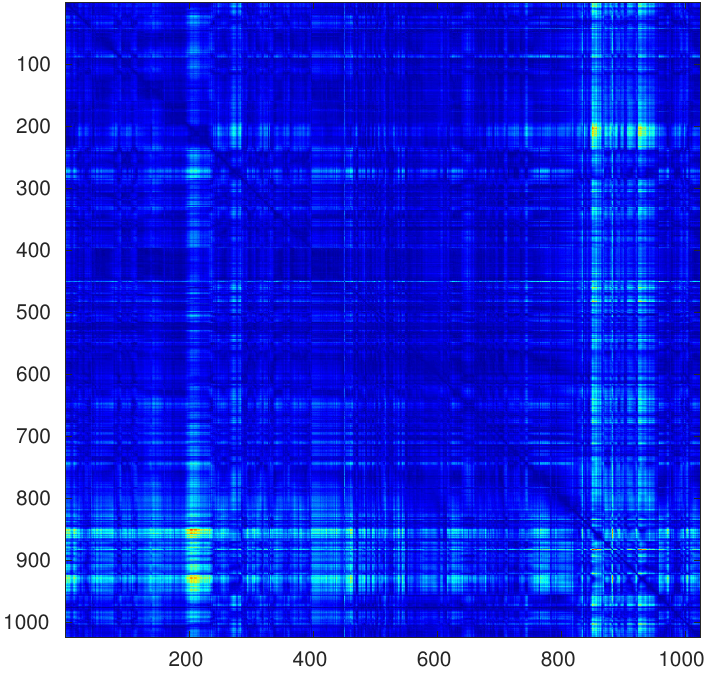}
\includegraphics[height=4cm]{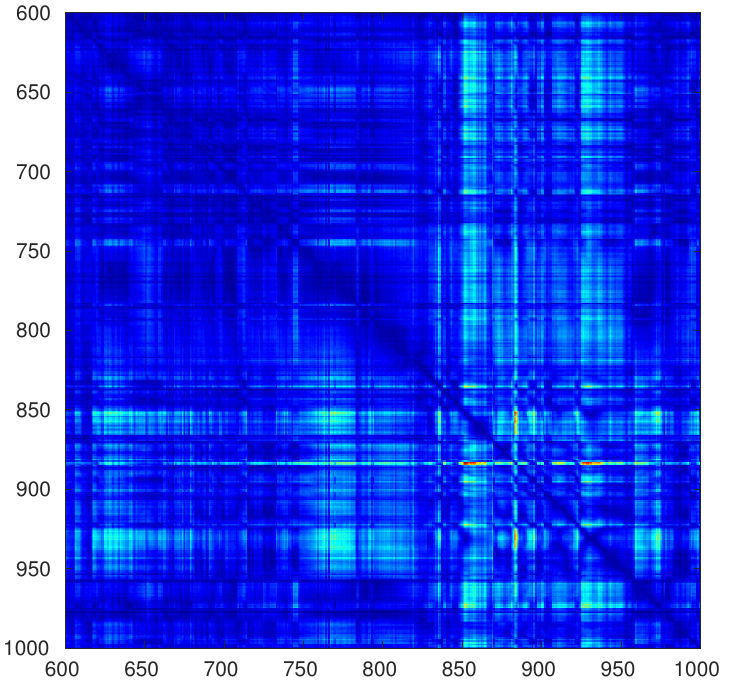}
\includegraphics[height=4cm]{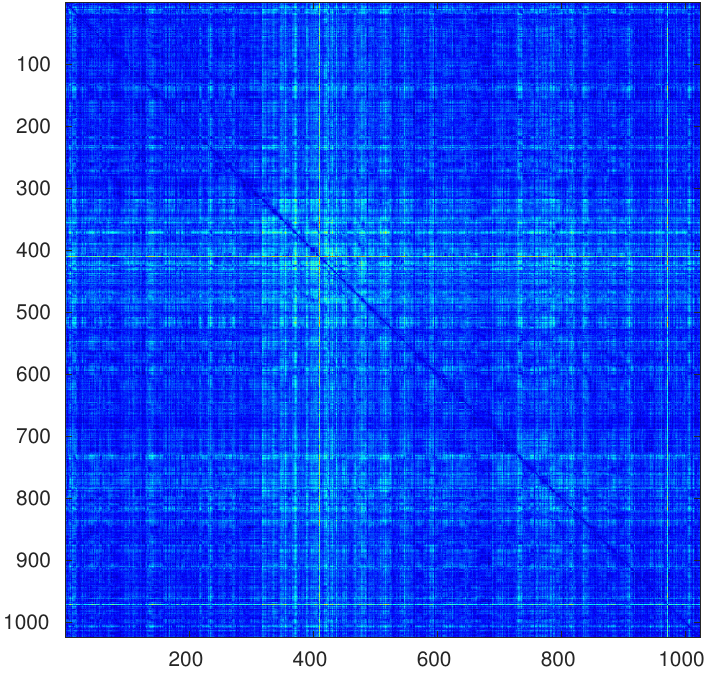}
\includegraphics[height=4cm]{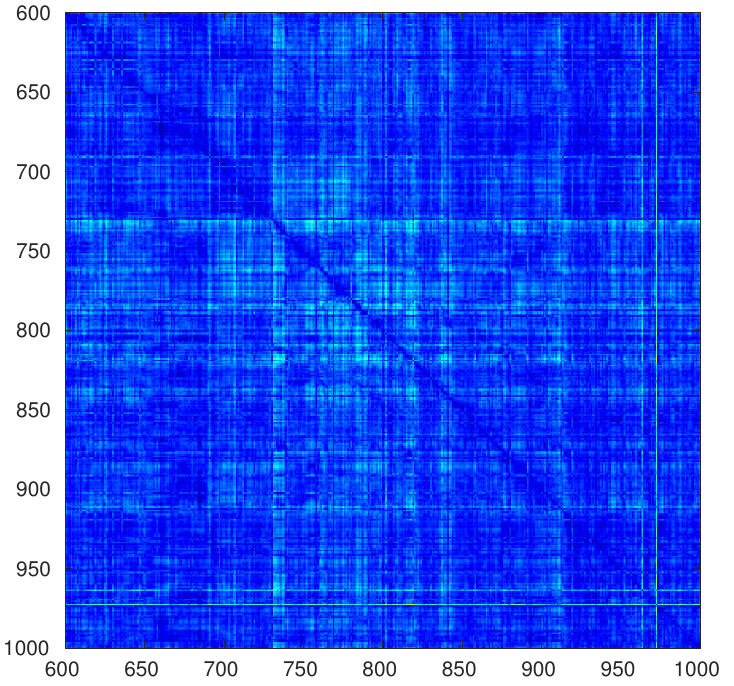}
}
\qquad
\subfloat[Evolution of the distances in feature space using different sequence length: (left to right) sequence length 20, 50, 100, 200. As the sequence gets longer, the matches become more distinctive: dominant diagonal and larger off-diagonal values.]{
\includegraphics[height=4cm]{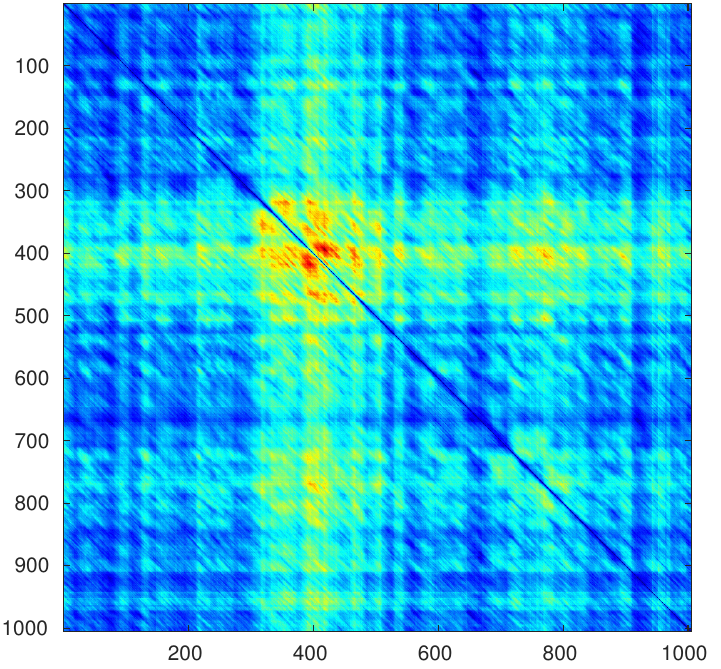}
\includegraphics[height=4cm]{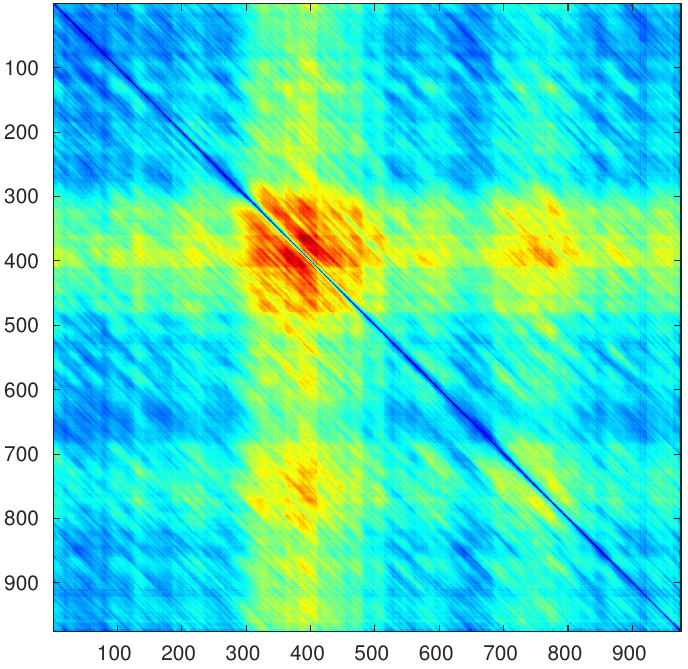}
\includegraphics[height=4cm]{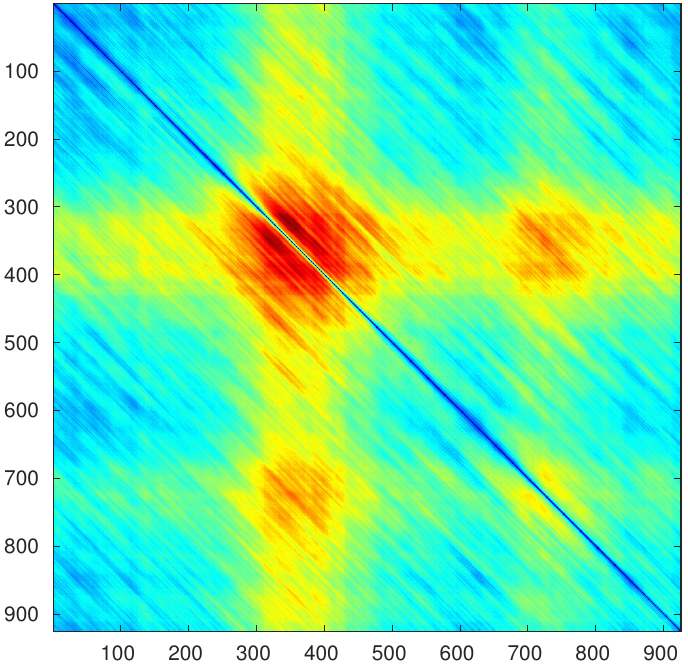}
\includegraphics[height=4cm]{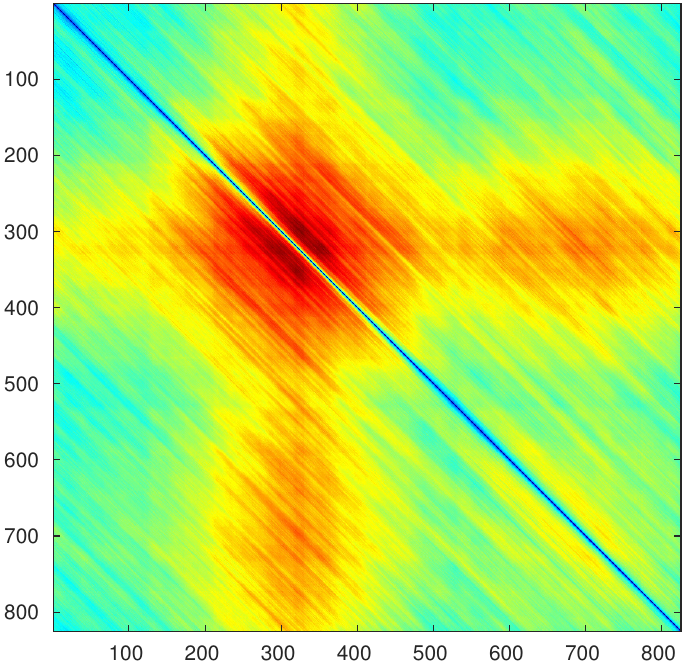}
}
\caption{Comparison of image-space (pixel-wise differences) to the feature space learned in the discriminators.}
\label{fig:dis_A}
\end{figure*}

\section{EXPERIMENTS}
\label{sec:experiments}
We demonstrate  the performance of our system using the Norland\footnote{http://nrkbeta.no/2013/01/15/nordlandsbanen-minute-by-minute-season-by-season/} dataset. 
The dataset has been recorded on board a train in Norway, 
over four-different seasons: summer, winter, autumn and spring.
The footage consists of about 10 hours of video under each weather condition which has been synchronized using GPS information. 
We extract image frames from the video at 2Hz leading to about 70 thousand images per weather condition. 
The dataset is split in half where the first half serves as the test set and the second half as the training set. 
For training, we further divide the training set into half again and use the first half of summer and the later half of the
winter images as training samples.
This is done to avoid any actual pairs being seen by the network. 
The training set finally contains about 17000 summer and as many winter images. 
All images are rescaled to a size of 64$\times$64. 
We train on color images instead of grayscale as it contains more information about the weather conditions. 
All experiment were carried out on a commodity GPU (GeForce GTX 980).
At test time, translation and feature generation takes less than 10ms per image on the GPU.

For the test set, we remove images where there is no apparent motion using a pixel-wise difference between images and use the first 1024 image pairs from summer and winter sequences,
allowing us to know the ground-truth correspondences.
In all the experiments, the output of the first fully connected layer in the discriminators is used as the feature descriptor.
All distance computation is done using cosine distances, unless mentioned otherwise.

\subsection{Learned Features}

We investigate if the distances in the learned feature space correspond to visual similarity,
that is, images close to each other in appearance, should have small distances between them. 
Using the discriminator is the summer domain, $\mathbf{D}_{S}$, we extract features from the first fully-connected layer and 
compute cosine distances between regularly sampled summer image. 
We randomly select some of theses and show their nearest-neighbours in the feature space (Fig.\ref{fig:similarImages}). 
The top row contains images where the train goes through tunnels and hence nothing can be seen.
It can be seen that indeed similar images are close to each other in this space.

We further investigate if these features are more suitable for place recognition across domains compared to pixel differences.
For this, using the appropriate generator, we translate winter images ($\mathbf{x}_{W}$) to get the corresponding summer images ($\mathbf{x}_{W \to S}$), 
for which a match is to be searched in the sequence of original summer images ($\mathbf{x}_{S}$).
Since the translation has put both images in the same domain (i.e. summer) and the correct match would come from a similar looking place,
we can compute the pixel-wise differences between them, which should be minimum for the correct match.
For each $\mathbf{x}_{W \to S,i}$, we calculate the cosine distance to all the summer images $\mathbf{x}_{S,j}$,
by treating the images as unit-vectors, leading to a distance matrix where each element (i,j) contains the distance 
between $\mathbf{x}_{W \to S,i}$ and $\mathbf{x}_{S,j}$.
Each row and column of this resulting distance matrix is normalized by the square-root of the diagonal entry at that row and column.
This normalization allows us to see the strength of the candidate matches in relation to the correct match (which is along the diagonal).
The resulting distances can be seen in Fig.\ref{fig:dis_A}(a)(1-2), 
which shows the normalized distance matrix along with a zoom-in on the region (600,1000). 
Higher distance are mapped as red. 

On the other hand, the discriminators in each domain learn features that are useful for classification,
thus can be utilized for the task of place recognition. 
For each $\mathbf{x}_{W \to S,i}$ and $\mathbf{x}_{S,j}$, we use the discriminator $\mathbf{D}_{S}$ to extract features 
from the first fully connected layer. 
We compute cosine distances between $\mathbf{x}_{W \to S,i}$ and $\mathbf{x}_{S,j}$, for each image pair.
The resulting distance matrix is normalized as before and given in Fig. \ref{fig:dis_A}(a)(3-4). 
It can be seen that distances in this space are much more meaningful compared to those in the image-space. 
This demonstrates that these learned features allow for more meaningful distances compared for pixel-differences in the image space.

A single feature, even in this learned space, may not be able to localize with sufficient accuracy. 
Instead, we can use sequences \cite{seqSLAM} of features to make localization more robust.
A simple idea to generate a feature for sequence of images is to  
concatenate the corresponding features from set of images, that is,
for a sequence length $n$ starting at time $t$, the feature is constructed by:
$\mathbf{f}_{t-n+1:t} = [~\mathbf{f}_{t-n+1}^{\mathbf{T}}~\mathbf{f}_{t-n}^{\mathbf{T}}~\hdots~\mathbf{f}_t^{\mathbf{T}}~]^{\mathbf{T}}$.

The effect of sequence lengths on distances in the feature space can be seen in Fig. \ref{fig:dis_A}(b),
which shows the normalized distance matrix for various sequence lengths.
The longer the sequences, the more distinctive the matches becomes and distances to other sequences becomes greater.

\subsection{Place recognition across summer and winter}
\begin{figure}
\centering
\subfloat[Precision Recall curves for nearest-neighbour matching in feature space. 
Longer sequences lead to better place recognition performance as features become more distinctive.
The highest precision is 99.8\% as increasing the sequence length prevents us from matching images at the start of the sequence.
Similarly for longer sequence length, highest recall is less than 100\%.]{
\includegraphics[width=.99\columnwidth]{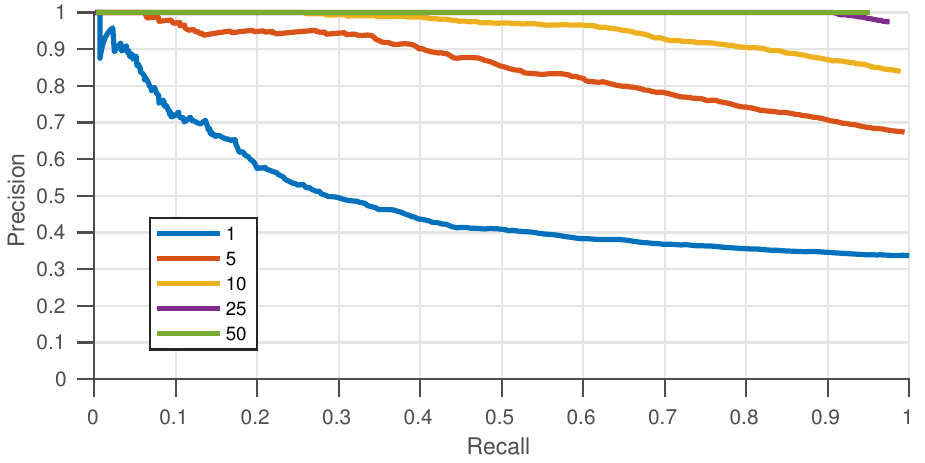}
}
\qquad
\subfloat[Precision Recall curves for summer and generate-summer images using SeqSLAM \cite{seqSLAM}. 
Generated-summer images are closer to summer images compared to winter image, leading to boost in performance for SeqSLAM.]{
\includegraphics[width=.99\columnwidth]{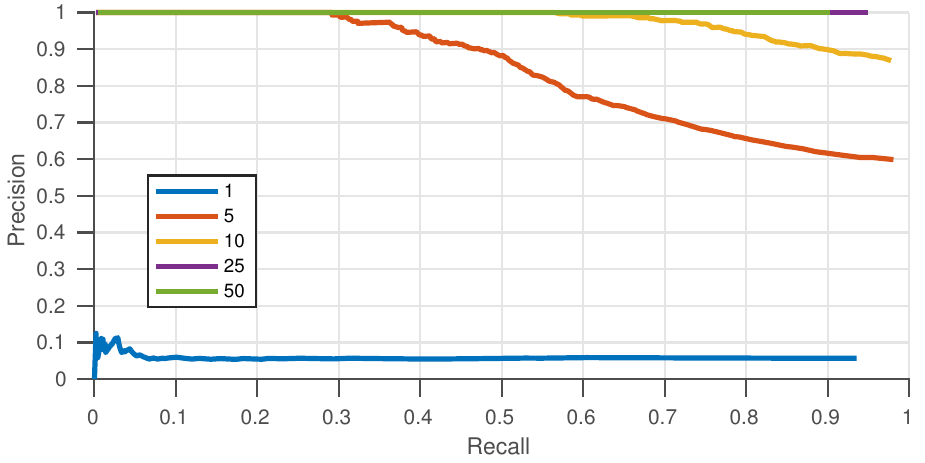}
}
\qquad
\subfloat[Precision Recall curves for summer and winter images using SeqSLAM]{
\includegraphics[width=.99\columnwidth]{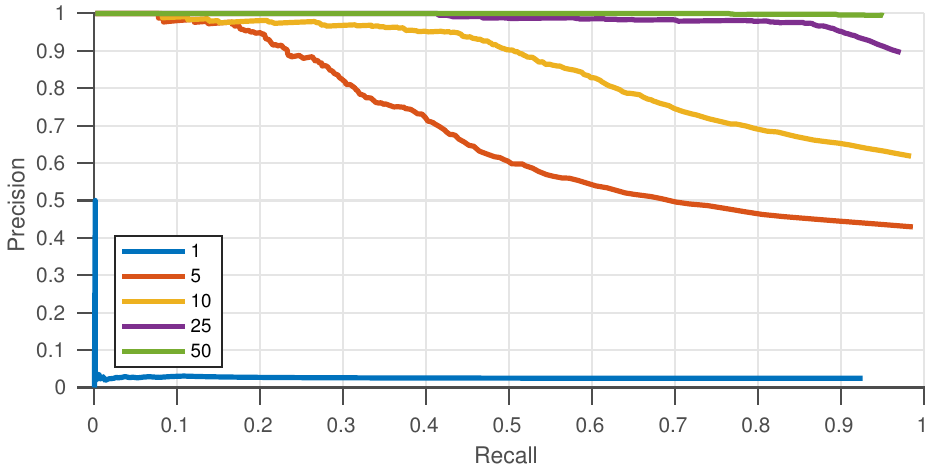}
}
\caption{Precision Recall curve for 1024 images from the Norland sequence. Each color corresponds to a different sequence length.
}
\label{fig:results}
\end{figure}

We perform a set of experiments that demonstrate the use of the learned features as well as the generated images for place recognition.
In particular, we use the concatenated discriminator feature vector, 
$\mathbf{f}_{t-n+1:t}$ directly to find the nearest-neighbour match in the feature space.

For nearest-neighbour, we threshold the maximum distance to generate precision-recall curves.
The results for various sequence length are given in Fig. \ref{fig:results}.
This experiment shows that learned features, along with the generator that translates from winter to summer, can
be effectively used for place recognition under severe illumination changes. 
significantly outperforming vanilla SeqSLAM (Fig. \ref{fig:results}(c)).

Additionally, we use the generated-summer image sequences in a traditional place recognition algorithm \cite{seqSLAM}. 
The performance (Fig. \ref{fig:results}(b)) improves compared to nearest-neighbour matching in feature space in all cases expect when the sequence length is 1.
This means that the learned features are better as discriminating between images, however, 
concatenating them is a very simple idea and more sophisticated algorithms can be designed.
 
Finally, we compare the performance of SeqSLAM on the original winter and summer images and the results are reported in Fig. \ref{fig:results}(c).
It can be seen that place recognition performance improves significantly when using generated-summer images, 
as they are perceptually closer to the original summer images, 
leading to a boost in performance for SeqSLAM.

\subsection{Conclusions and Future Work}
\begin{figure}
\centering
\includegraphics[width=.95\columnwidth]{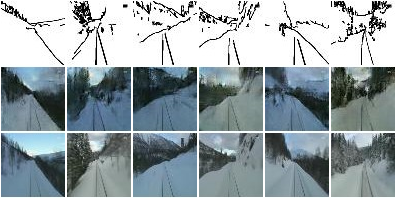}
\caption{Translation between edges and winter : The edges were extracted from winter images using Canny edge detector
and serve as the second domain to learn. The first row shows the edge, the second is the winter image generated from the
edges and last row shows the corresponding true winter image.}
\label{fig:winterEdges}
\vspace*{-5mm}
\end{figure}

In this work, we explored the possibility of using GANs for the task of image translation and subsequently using the
features spaces learned in the discriminator for place recognition. 
We show that these features encode visually similar images close in the space
and can be used for place recognition using generated images in the domain of interest. 
In addition to using the learned features, the generated images can be used in a traditional place recognition system to get an improvement in performance.

While it is interesting to be able to translate between a pair of domains, 
in order to close loops between many different conditions, all pairs of possible translations have to be learned,
which is not feasible. 
This thought has motivated us to look into domains that can trivially be computed from each domain, such as structure. 
We can represent structure using lines in the image and learn how to associate an image to its edges, some initial
results are shown in Fig. \ref{fig:winterEdges}. 
Such a setup would allow us to learn an embedding for edges in the
image, which can then be used to match edges between different weather conditions.  



\section*{ACKNOWLEDGMENT}
This work was supported by the Australian Research Council 
Centre of Excellence for Robotic Vision CE140100016, 
and through a Laureate Fellowship FL130100102 to IR. 



\end{document}